\documentclass[11pt]{article}
\font\tenmath=msbm10
\font\sevenmath=msbm7
\font\fivemath=msbm5
\newfam\mathfam \textfont\mathfam=\tenmath
\scriptfont\mathfam=\sevenmath \scriptscriptfont\mathfam=\fivemath

\usepackage{amsmath}
\usepackage{amssymb}
\usepackage{setspace}
\usepackage{amsthm}
\usepackage{natbib}

\usepackage{hypernat}
\usepackage{graphics}
\usepackage{graphicx}

\usepackage{amsmath}

\usepackage[colorlinks,citecolor=blue,urlcolor=blue]{hyperref}

\newcommand{\cA}{\mathcal{A}}
\newcommand{\cE}{\mathcal{E}}
\newcommand{\cF}{\mathcal{F}}
\newcommand{\cH}{\mathcal{H}}
\newcommand{\cO}{\mathcal{O}}

\newcommand{\cX}{\mathcal{X}}

\newcommand{\R}{{\rm I}\kern-0.18em{\rm R}}
\newcommand{\p}{{\rm I}\kern-0.18em{\rm P}}
\newcommand{\E}{{\rm I}\kern-0.18em{\rm E}}
\newcommand{\1}{{\rm 1}\kern-0.24em{\rm I}}

\newcommand{\field}[1]{\mathbb{#1}}

\newcommand{\hh}{\textsf{h}}

\newcommand{\card}{\mathop{\mathrm{card}}}
\newcommand{\argmin}{\mathop{\mathrm{argmin}}}

\newcommand{\ud}{\mathrm{d}}
\newcommand{\epr}{\hfill $\square$\vspace{0.5cm}\par}

\newcommand{\eps}{\varepsilon}

\newtheorem{TH1}{Theorem}[section]
\newtheorem{prop}{Proposition}[section]
\newtheorem{cor}{Corollary}[section]
\newtheorem{lem}{Lemma}[section]
\newtheorem{hyp}{Assumption}
\newtheorem{defin}{Definition}[section]

\newcommand{\HC}{{\mathcal{H}^{\mathrm{conv}}}}
\newcommand{\Rm}{R_\varphi^{{\scriptscriptstyle -}}}
\newcommand{\Rp}{R_\varphi^{{\scriptscriptstyle +}}}
\newcommand{\hRm}{\hat{R}_\varphi^{{\scriptscriptstyle -}}}
\newcommand{\hRp}{\hat{R}_\varphi^{{\scriptscriptstyle +}}}
\newcommand{\Pm}{P^{{\scriptscriptstyle -}}}
\newcommand{\Pp}{P^{{\scriptscriptstyle +}}}
\newcommand{\Xm}{X^{{\scriptscriptstyle -}}}
\newcommand{\Xp}{X^{{\scriptscriptstyle +}}}
\newcommand{\Nm}{N^{{\scriptscriptstyle -}}}
\newcommand{\Np}{N^{{\scriptscriptstyle +}}}
\newcommand{\snm}{n^{{\scriptscriptstyle -}}}
\newcommand{\snp}{n^{{\scriptscriptstyle +}}}

\begin{document}
\title{\bf Neyman-Pearson classification, convexity and stochastic constraints}
\author{
{\sc Philippe Rigollet}\thanks{Supported by NSF grant DMS-0906424} \quad  {\normalsize and} \quad {\sc Xin Tong}\\
\normalsize Princeton University\\
\normalsize \{rigollet, xtong\}@princeton.edu}
\date{\normalsize \today}
\maketitle

\maketitle

\begin{abstract}
Motivated by problems of anomaly detection, this paper implements the Neyman-Pearson paradigm to deal with asymmetric errors in binary classification with a convex loss. Given a finite collection of classifiers, we combine them and obtain a new classifier  that satisfies simultaneously the two following properties with high probability: (i) its probability of type~I error is below a pre-specified level and (ii), it has probability of type~II error close to the minimum possible. The proposed classifier is obtained by solving an optimization problem with an empirical objective and an empirical constraint.  New techniques to handle such problems are developed and have consequences on chance constrained programming.
\end{abstract}

%



\textbf{keywords}:  binary classification, Neyman-Pearson paradigm, anomaly detection, stochastic constraint, convexity, empirical risk minimization, chance constrained optimization.

\section{Introduction}
The Neyman-Pearson (NP) paradigm in statistical learning extends the objective of classical binary classification in that, while the latter focuses on minimizing classification error that is a weighted sum of type~I and type~II errors, the former minimizes type~II error with an upper bound $\alpha$ on type~I error. With slight abuse of language, in verbal discussion we do not distinguish type~I/II error from probability of type~I/II error.

For learning with the NP paradigm, it is essential to avoid one kind of error at the expense of the other. As an illustration, consider the following problem in medical diagnosis: failing to detect a malignant tumor has far more severe consequences than flagging a benign tumor. Other scenarios include spam filtering, machine monitoring, target recognition, etc.

In the learning context, as true errors are inaccessible, we cannot enforce almost surely the desired upper bound for type~I error. The best we can hope is that a data dependent classifier has type~I error bounded with high probability.  Henceforth,  there are two goals in this project. The first is to design a learning procedure so that type~I error of the learned classifier $\hat{f}$ is upper bounded by a pre-specified level with pre-specified high probability; the second is to show that $\hat{f}$ has good performance bounds for excess type ~II error.

This paper is organized as follows. In Section~\ref{SEC:classif}, the classical setup for binary classification is reviewed and the main notation is introduced. A parallel between binary classification and statistical hypothesis testing is drawn in Section~\ref{SEC:NPclassif} with emphasis on the NP paradigm in both frameworks. The main propositions, theorems and their proofs are stated in Section~\ref{SEC:main} while secondary, technical results are relegated to the Appendix. Finally, Section~\ref{SEC:CCP} illustrates an application of our results to chance constrained optimization.

In the rest of the paper, we denote by $x_j$ the $j$-th coordinate of a vector $x \in \R^d$.
\section{Binary classification}
\label{SEC:classif}
\subsection{Classification risk and classifiers}
Let $(X, Y)$ be a random couple where $X \in \cX \subset \R^d$ is a vector of covariates and $Y \in \{-1,1\}$ is  a label that indicates to which class $X$ belongs.
A \emph{classifier} $h$ is a mapping $h: \mathcal{X}\rightarrow [-1,1]$ whose sign returns the predicted class given $X$.  An error occurs when $-h(X)Y\geq0$ and it is therefore natural to define the classification loss by $\1(-h(X)Y\geq0)$, where $\1(\cdot)$ denotes the indicator function.

The expectation of the classification loss with respect to the joint distribution of $(X,Y)$ is called \emph{(classification) risk} and is defined by
\begin{align*}
R(h)=\field{P}\left(-h(X)Y\geq0\right).
\end{align*}
Clearly, the indicator function is not convex and for computation, a common practice is to replace it by a convex surrogate~\citep[see, e.g.][and references therein]{BarJorMcA06}.

To this end, we rewrite the risk function as
\begin{equation}\label{surrogate}
R(h)
= \E [\varphi(-h(X)Y)],
\end{equation}
where $\varphi(z)=\1\left(z\geq0\right)$. Convex relaxation can be achieved by simply replacing the indicator function by a convex surrogate.
\begin{defin}
A function $\varphi: [-1, 1] \to \mathbb{R}^+$ is called a \emph{convex surrogate} if it is non-decreasing, continuous and convex and if $\varphi(0)=1$.
\end{defin}
Commonly used examples of convex surrogates are the hinge loss $\varphi(x)=(1+x)_+$, the logit loss $\varphi(x)=\log_2 (1+e^x)$ and the exponential loss $\varphi(x)=e^x$.

For a given choice of $\varphi$, define the $\varphi$-risk
\begin{align*}
R_{\varphi}(h)=\E [\varphi(-Yh(X))]\,.
\end{align*}
Hereafter, we assume that $\varphi$ is fixed and refer to $R_\varphi$ as the risk.
In our subsequent analysis, this convex relaxation will also be the ground to analyze a stochastic convex optimization problem subject to stochastic constraints. A general treatment of such problems can be found in Section~\ref{SEC:CCP}.

Because of overfitting, it is unreasonable to look for mappings minimizing empirical risk over all calssifiers.  Indeed, one could have a small empirical risk but a large true risk. Hence, we resort to regularization.  There are in general two ways to proceed.  The first is to restrict the candidate classifiers to a specific class $\mathcal{H}$, and the second is to change the objective function by, for example, adding a penalty term.  The two approaches can be combined, and sometimes are obviously equivalent.

In this paper, we pursue the first idea by defining the class of candidate classifiers as follows. Let $h_{1},\ldots, h_{M}, M\ge 2$ be a given collection of classifiers. In our setup, we allow $M$ to be large. In particular, our results remain asymptotically meaningful as long as $M=o(e^n)$. Such classifiers are usually called base classifiers and can be constructed in a very naive manner. Typical examples include decision stumps or small trees. While the $h_j$'s may have no satisfactory classifying power individually, for over two decades, boosting type of algorithms have successfully exploited the idea that a suitable weighted majority vote among these classifiers may result in low classification risk~\citep{Sch90}. Consequently, we restrict our search for classifiers to the set of functions consisting of convex combinations of the $h_j$'s:
\begin{align*}
\HC=\{\textsf{h}_{\lambda}=\sum_{j=1}^M\lambda_{j}h_{j},\lambda\in \Lambda\},
\end{align*}
where $\Lambda$ denotes the flat simplex of $\R^M$ and is defined by $\Lambda=\{\lambda\in \R^M \,:\, \lambda_j \ge 0,  \sum_{j=1}^M\lambda_{j}=1\}$. In effect, classification rules given by the sign of $h \in \HC$ are exactly the set of rules produced by the weighted majority votes among the base classifiers $h_1, \ldots, h_M$.

By restricting our search to classifiers in $\HC$, the best attainable $\varphi$-risk is called \emph{oracle risk} and is abusively denoted by $R_\varphi(\HC)$. As a result, we have $R_\varphi(h)\ge R_\varphi(\HC)$ for any $h \in \HC$ and a natural measure of performance for a classifier $h \in \HC$ is given by its excess risk defined by $R_\varphi(h)-R_\varphi(\HC)$.

The excess risk of a data driven classifier $h_n$ is a random quantity and we are interested in bounding it with high probability. Formally, the statistical goal of binary classification is to construct a classifier $h_n$ such that the oracle inequality
\begin{align}\label{OI_binary}
R_\varphi(h_n)
\leq R_\varphi(h_{\HC})+ \Delta_n(\HC, \delta)\,
\end{align}
holds with probability $1-\delta$, where $\Delta_n(\cdot, \cdot)$ should be as small as possible.

In the scope of this paper, we focus on candidate classifiers in the class $\cH^{\rm conv}$. Some of the following  results such as Theorem~\ref{TH:typeI} can be extended to more general classes of classifiers with known complexity such as classes with bounded VC-dimension, as for example in~\citet{CanHowHus02}. However, our main argument for bounding type~II error relies on Proposition~\ref{PROP:stochastic_constraint} which, in turn, depends heavily on the convexity of the problem, and it is not clear how it can be extended to more general classes of classifiers.

\subsection{The Neyman-Pearson paradigm}

In classical binary classification, the risk function can be expressed as a convex combination of type~I error $R^{{\scriptscriptstyle -}}(h)=\p \left(-Yh(X)\geq0|Y=-1\right)$ and  of type~II error $R^{{\scriptscriptstyle +}}(h)=\p \left(-Yh(X)\geq0|Y=1\right)$:
\begin{equation*}
R(h)=\p(Y=-1)R^{{\scriptscriptstyle -}}(h)+\p(Y=1)R^{{\scriptscriptstyle +}}(h).
\end{equation*}
More generally, we can define the $\varphi$-type~I and  $\varphi$-type~II errors respectively by
$$
\Rm(h)=\E \left[\varphi(-Yh(X))|Y=-1\right]\qquad \mathrm{and} \qquad \Rp(h)=\E \left[\varphi(-Yh(X))|Y=1\right].
$$

Following the NP paradigm, for a given class $\cH$ of classifiers, we seek to solve the constrained minimization problem:
\begin{align}\label{EQ:optim}
\min_{\substack{h\in\mathcal{H}\\ \Rm(h)\leq\alpha}}\Rp(h),
\end{align}
where $\alpha\in(0,1)$, the significance level,  is a constant specified by the user.

NP classification is closely related to the NP approach to statistical hypothesis testing.  We now recall a few key concepts about the latter. Many classical works have addressed the theory of statistical hypothesis testing, in particular \citet{LehRom05} provides a thorough treatment of the subject.

Statistical hypothesis testing bears strong resemblance with binary classification if we assume the following  model. Let $\Pm$ and $\Pp$ be two probability distributions on $\cX \subset \R^d$. Let $p \in (0,1)$ and assume that $Y$ is a random variable defined by
$$
Y=\left\{
\begin{array}{ll}
1 & \textrm{with probability } p\,,\\
-1 & \textrm{with probability } 1-p\,.
\end{array}\right.
$$
Assume further that the conditional distribution of $X$ given $Y$ is given by $P^Y$. Given such a model, the goal of statistical hypothesis testing is to determine whether $X$ was generated from $\Pm$ or $\Pp$. To that end, we construct a test $\phi:\cX \to [0,1]$ and the conclusion of the test based on $\phi$ is that $X$ is generated from $\Pp$ with probability $\phi(X)$ and from $\Pm$ with probability $1-\phi(X)$. Note that randomness here comes from an exogenous randomization process such as flipping a biased coin. Two kinds of errors arise: type~I error occurs when rejecting $\Pm$ when it is true, and type~II error occurs when accepting $\Pm$ when it is false. The Neyman-Pearson paradigm in hypothesis testing amounts to choosing $\phi$ that solves the following constrained optimization problem
$$
\begin{array}{rl}
\text{maximize } &\E[\phi(X)|Y=1]\,,\\
\textrm{subject to } & \E[\phi(X)|Y=-1]\leq\alpha\,,
\end{array}
$$
where $\alpha \in (0,1)$ is the significance level of the test. In other words, we specify a significance level $\alpha$ on type~I error, and minimize  type~II error. We call a solution to this problem \emph{a most powerful test} of level $\alpha$. The Neyman-Pearson Lemma gives mild sufficient conditions for the existence of such a test.

\begin{TH1}[Neyman-Pearson Lemma]\label{TH:NP}
Let $\Pm$ and $\Pp$ be probability distributions possessing densities $p^{{\scriptscriptstyle -}}
$ and $p^{{\scriptscriptstyle +}}$ respectively with respect to some measure $\mu$. Let $\varphi_k(x)=\1\left(L(x)\geq k\right)$, where the likelihood ratio $L(x)=p^{{\scriptscriptstyle +}}(x)/p^{{\scriptscriptstyle -}}(x)$ and $k$ is such that
$\Pm(L(X)>k)\leq\alpha$ and $\Pm(L(X)\geq k)\geq\alpha$.  Then,
\begin{itemize}
\item  $\varphi_{k}$ is a level $\alpha=\E\left[\varphi_k(X)|Y=-1\right]$ most powerful test.
\item For a given level $\alpha$, the most powerful test of level $\alpha$ is defined by
\begin{equation*}
\phi(X)=\left\{
 \begin{array}{lll}
     1 & \text{if } & L(X)>k\\
     0 & \text{if } & L(X)<k\\
     \frac{\alpha-\Pm(L(X)>k)}{\Pm(L(X)=k)} & \text{if } & L(X)=k
   \end{array}      \right.
\end{equation*}
\end{itemize}
\end{TH1}

Notice that in the learning framework,  $\phi$ cannot be computed since it requires the knowledge of the likelihood ratio and of the distributions $\Pm$ and $\Pp$. Therefore, it remains merely a theoretical propositions. Nevertheless, the result motivates the NP paradigm pursued here.

\section{Neyman-Pearson classification via convex optimization}
\label{SEC:NPclassif}
\setcounter{equation}{0}

Recall that in NP classification, the goal is to solve the problem \eqref{EQ:optim}. This cannot be done directly as conditional distributions $\Pm$ and $\Pp$, and hence $\Rm$ and $\Rp$, are unknown. In statistical applications, information about these distributions is available through two i.i.d. samples $\Xm_1, \ldots, \Xm_{\snm}$, $\snm\ge 1$ and $\Xp_1, \ldots, \Xp_{\snp}$, $\snp \ge 1$,  where $\Xm_i \sim \Pm, i=1, \ldots, \snm$ and $\Xp_i\sim \Pp, i=1, \ldots, \snp$. We do not assume that the two samples $(\Xm_1, \ldots, \Xm_{\snm})$ and $(\Xp_1, \ldots, \Xp_{\snp})$ are mutually independent. Presently the sample sizes $\snm$ and $\snp$ are assumed to be deterministic and will appear in the subsequent finite sample bounds. A different sampling scheme, where these quantities are random, is investigated in subsection~\ref{sub:sampling}.

\subsection{Previous results and new input}

While the binary classification problem has been extensively studied, theoretical proposition on how to implement the NP paradigm remains scarce. To the best of our knowledge, \citet{CanHowHus02} initiated the theoretical treatment of the NP classification paradigm and an early empirical study can be found in~\citet{CasChe03}. The framework of \citet{CanHowHus02} is the following. Fix a constant $\varepsilon_0>0$ and let $\mathcal{H}$ be a given set of classifiers with finite VC dimension. They study a procedure that consists of solving the following relaxed empirical optimization problem
\begin{equation}\label{EQ:Cannon1}
\min_{\substack{h\in\mathcal{H}\\ \hat{R}^{{\scriptscriptstyle -}}(h)\leq\alpha+\varepsilon_0/2}}\hat{R}^{{\scriptscriptstyle +}}(h),
\end{equation}
where
$$
\hat{R}^{{\scriptscriptstyle -}}(h)=\frac{1}{\snm}\sum_{i=1}^{\snm} \1(h(\Xm_i) \ge 0)\,, \quad \mathrm{and}\quad \hat{R}^{{\scriptscriptstyle +}} (h)=\frac{1}{\snp}\sum_{i=1}^{\snp} \1(h(\Xm_i) \le 0)\,
$$
denote the empirical type~I and empirical type~II errors respectively. Let $\hat{h}$ be a solution to \eqref{EQ:Cannon1}. Denote by $h^{*}$ a solution to the original Neyman-Pearson optimization problem:
\begin{equation}
\label{EQ:NPoriginal}
h^* \in \argmin_{\substack{h\in\mathcal{H}\\ R^{{\scriptscriptstyle -}}(h)\leq\alpha}}R^{{\scriptscriptstyle +}}(h)\,,
\end{equation}
The main result of~\citet{CanHowHus02} states that, simultaneously with high probability, the type~II error $R^{{\scriptscriptstyle +}}(\hat h)$ is bounded from above by $R^{{\scriptscriptstyle +}}(h^{*})+\varepsilon_1$, for some $\varepsilon_1>0$ and the type~I error of $\hat{h}$ is bounded from above by $\alpha+\epsilon_0$. In a later paper, \citet{CanHowHus03} considers problem \eqref{EQ:Cannon1} for a data-dependent family of classifiers $\mathcal{H}$, and bound estimation errors accordingly. Several results for traditional statistical learning such as PAC bounds or oracle inequalities have been studied in \citet{Sco05} and \citet{ScoNow05} in the same framework as the one laid down by~\citet{CanHowHus02}. A noteworthy departure from this setup is \citet{Sco07} where sensible performance measures for NP classification that go beyond analyzing separately two kinds of errors are introduced. Furthermore,  \citet{BlaLeeSco09} develops a general solution to semi-supervised novelty detection by reducing it to NP classification.
Recently, \citet{HanCheSun08} transposed several results of \citet{CanHowHus02} and \citet{ScoNow05} to NP classification with convex loss.

The present work departs from previous literature in our treatment of type~I error. As a matter of fact, the classifiers in all the papers mentioned above can only ensure that  $\p(R^{{\scriptscriptstyle -}}(\hat{h})>\alpha+\varepsilon_0)$ is small, for some $\epsilon_0>0$.  However, it is our primary interest to make sure that $R^{{\scriptscriptstyle -}}(\hat{h})\le \alpha$ with high probability, following the original principle of the Neyman-Pearson paradigm that type~I error should be controlled by a pre-specified level $\alpha$. As will be illustrated, to control $\p(R^{{\scriptscriptstyle -}}(\hat{h})>\alpha)$, it is necessary to have $\hat{h}$ be a solution to some program with a strengthened constraint on empirical type~I error. If our concern is only on type~I error, we can just do so.  However, we also want to control excess type~II error simultaneously.

The difficulty was foreseen in the seminal paper \citet{CanHowHus02}, where it is claimed without justification that if we use $\alpha'<\alpha$ for the empirical program, ``it seems unlikely that we can control the estimation  error $R^{{\scriptscriptstyle +}}(\hat{h})-R^{{\scriptscriptstyle +}}(h^*)$ in a distribution independent way". The following proposition confirms this opinion in a certain sense.

Fix $\alpha \in (0,1), \snm\ge1, \snp\ge 1$ and $\alpha'<\alpha$. Let $\hat h(\alpha')$ be the classifier defined as any solution of the following optimization problem:
$$
\min_{\substack{h\in\mathcal{H}\\ \hat{R}^{{\scriptscriptstyle -}}(h)\leq\alpha'}}\hat{R}^{{\scriptscriptstyle +}}(h)\,.
$$
The following negative result holds not only for this estimator but also for the oracle $h^*(\alpha')$ defined as the solution of
$$
\min_{\substack{h\in\mathcal{H}\\ {R}^{{\scriptscriptstyle -}}(h)\leq\alpha'}}{R}^{{\scriptscriptstyle +}}(h)\,.
$$
Note that $h^*(\alpha')$ is not a classifier but only a pseudo-classifier since it depends on the unknown distribution of the data.
\begin{prop}
\label{PROP:cannon}
There exist base classifiers $h_1, h_2$ and a probability distribution for $(X, Y)$ for which, regardless of the sample sizes $\snm$ and $\snp$,  any pseudo-classifier $h\in [h_1, h_2]$ such that ${R}^{{\scriptscriptstyle -}}(h)<\alpha$, it holds
$$
{R}^{{\scriptscriptstyle +}}(h)-\min_{\lambda\in [0,1]}{R}^{{\scriptscriptstyle +}}(\lambda h_1 + (1-\lambda)h_2) \ge \alpha >0\,.
$$
In particular, the excess type~II risk of $h^*(\alpha-\eps_{\snm}), \,\eps_{\snm}>0$ does not converge to zero as sample sizes increase even if $\eps_{\snm}\to 0$.
Moreover, when $\alpha\leq 1/2$ for any (pseudo-)classifier $h\in [h_1, h_2]$ such that ${\hat R}^{{\scriptscriptstyle -}}(h)<\alpha$, it holds
\begin{equation}
\label{EQ:prop_cannon_2}
{R}^{{\scriptscriptstyle +}}(h)-\min_{\lambda\in [0,1]}{R}^{{\scriptscriptstyle +}}(\lambda h_1 + (1-\lambda)h_2) \ge \alpha >0\,.
\end{equation}
with probability at least $\alpha \wedge 1/4$. In particular, the excess type~II risk of $\hat h(\alpha-\eps_{\snm}), \,\eps_{\snm}>0$ does not converge to zero with positive probability, as sample sizes increase even if $\eps_{\snm}\to 0$.
\end{prop}


The proof of this result is postponed to the appendix. The fact that the oracle $h^*(\alpha-\eps_{\snm})$ satisfies the lower bound indicates that the problem comes from using a strengthened constraint. Note that the condition $\alpha\le 1/2$ is purely technical and can be removed. Nevertheless, it is always the case in practice that $\alpha\le 1/2$.

In view of this negative result, it seems that our rightful insist on type~I error does not go well with the ambition to control type~II error simultaneously.  To overcome this dilemma, we resort to a continuous convex surrogate as our loss function.  In particular, we design a modified version of empirical risk minimization method such that the data-driven classifier $\hat{h}$ has type~I error bounded by $\alpha$ with high probability.  Moreover, we consider here a class~$\cH$ that allows a different treatment of the empirical processes involved.

This new approach comes with new technical challenges which we summarize here. In the approach of~\citet{CanHowHus02} and of~\citet{ScoNow05}, the relaxed constraint on the type~I error is constructed such that the constraint $\hat{R}^{{\scriptscriptstyle -}}(h)\leq\alpha+\varepsilon_0/2$ on type~I error in~\eqref{EQ:Cannon1} is satisfied by $h^*$ (defined in \eqref{EQ:NPoriginal}) with high probability, and that this classifier accommodates excess type~II error well.  As a result, the control of type~II error mainly follows as a standard exercise to control suprema of empirical processes. This is not the case here; we have to develop methods to control the optimum value of a convex optimization problem under a stochastic constraint. Such methods have consequences not only in NP classification but also on chance constraint programming as explained in Section~\ref{SEC:CCP}.

\subsection{Convexified NP classifier}
\label{sub:defin}
To solve the problem of NP classification~\eqref{EQ:optim} where the distribution of the observations is unknown, we resort to empirical risk minimization. In view of the arguments presented in the previous subsection, we cannot simply replace the unknown true risk functions by their empirical counterparts. The treatment of the convex constraint should be done carefully and we proceed as follows.

For any classifier $h$ and a given convex surrogate $\varphi$, define $\hRm$ and $\hRp$ to be the empirical counterparts of $\Rm$ and $\Rp$ respectively by
$$
\hRm (h)=\frac{1}{\snm}\sum_{i=1}^{\snm} \varphi(h(\Xm_i))\,, \quad \text{and}\quad \hRp (h)=\frac{1}{\snp}\sum_{i=1}^{\snp} \varphi(-h(\Xp_i))\,.
$$

Moreover, for any $a>0$, let $\mathcal{H}^{\varphi, a}=\{h \in \HC \,:\, \Rm(h)\leq a\}$ be the set of classifiers in $\HC$ whose convexified type~I errors are bounded from above by $a$, and let $\mathcal{H}_{\snm}^{\varphi, a}= \{h \in \HC \,:\, \hRm(h)\leq a\}$ be the set of classifiers in $\HC$ whose empirical convexified type~I errors are bounded by $a$.
To make our analysis meaningful, we assume that $\mathcal{H}^{\varphi, \alpha} \neq\emptyset$.

We are now in a position to construct a classifier in~$\HC$ according to the Neyman-Pearson paradigm. For any $\tau>0$ such that $\tau \le \alpha\sqrt{\snm}$, define the convexified NP classifier $\tilde h^\tau$ as any classifier that solves the following optimization problem
\begin{equation}
\label{EQ:NPclassifier}
\min_{\substack{h\in\HC\\\hRm(h)\leq\alpha-\tau/\sqrt{\snm}}}\hRp(h)\,.
\end{equation}
Note that this problem consists of minimizing a convex function subject to a convex constraint and can therefore be solved by standard algorithms such as~\citep[see, e.g.,][and references therein]{BoyVan04}.

In the next section, we present a series of results on type~I and type~II errors of classifiers that are more general than $\tilde h^\tau$.

\section{Performance Bounds}
\label{SEC:main}
\setcounter{equation}{0}

\subsection{Control of type~I error}
The first challenge is to identify classifiers $h$ such that $\Rm(h) \le \alpha$ with high probability. This is done by enforcing its empirical counterpart $\hRm(h)$ be bounded from above by the quantity
\begin{equation*}
\alpha_\kappa=\alpha-\kappa/\sqrt{\snm},
\end{equation*}
for a proper choice of positive constant $\kappa$.

\begin{TH1}\label{TH:typeI}
Fix constants $\delta, \alpha\in(0,1), L>0$ and let $\varphi:[-1,1]\to \R^+$ be a given $L$-Lipschitz convex surrogate. Define
$$
\kappa=4\sqrt{2}L\sqrt{\log \left(\frac{2M}{\delta}\right)}\,.
$$
Then for any (random) classifier $h \in\HC$ that satisfies $\hRm(h)\leq \alpha_\kappa$, we have
$$
R^{{\scriptscriptstyle -}}(h) \le \Rm( h)\leq \alpha\,.
$$
with probability at least $ 1-\delta$. Equivalently
\begin{equation}
\label{EQ:inclusion_whp}
\p\left[\mathcal{H}_{\snm}^{\varphi, \alpha_\kappa }\subset \mathcal{H}^{\varphi, \alpha} \right]\ge 1-\delta\,.
\end{equation}
\end{TH1}

\subsection{Simultaneous control of the two errors}

Theorem~\ref{TH:typeI} guarantees that any classifier that satisfies the strengthened constraint on the empirical $\varphi$-type~I error will have $\varphi$-type~I error and true type~I error bounded from above by~$\alpha$. We now check that the constraint is not too strong so that the type~II error is overly deteriorated. Indeed, an extremely small $\alpha_\kappa$ would certainly ensure a good control of type~I error but would deteriorate significantly the best achievable type~II error. Below, we show not only that this is not the case for our approach but also that the convexified NP classifier $\tilde h^\tau$ defined in subsection~\ref{sub:defin} with $\tau=\alpha_\kappa$ suffers only a small degradation of its type~II error compared to the best achievable.
Analogues to classical binary classification, a desirable result is that  with high probability,
\begin{equation}\label{NP oracle}
\Rp(\tilde h^{\alpha_\kappa})-\min_{h \in \mathcal{H}^{\varphi, \alpha}}\Rp(h)
\leq \tilde{\Delta}_{n}(\mathcal{F}),
\end{equation}
where $\tilde{\Delta}_{n}(\mathcal{F})$ goes to $0$ as $n=\snm+\snp\rightarrow\infty$.

The following proposition is pivotal to our argument.
\begin{prop}
\label{PROP:stochastic_constraint}
Fix constant $\alpha\in(0,1)$ and let $\varphi:[-1,1]\to \R^+$ be a given continuous convex surrogate. Assume further that there exists $\nu_0>0$ such that  the set  of classifiers $\mathcal{H}^{\varphi, \alpha-\nu_0}$ is nonempty. Then, for any $\nu \in (0, \nu_0)$,
$$
\min_{h \in \cH^{\varphi, \alpha-\nu}} \Rp(h)-\min_{h \in \cH^{\varphi, \alpha}}\Rp(h)
\leq \varphi(1)\frac{\nu}{\nu_0-\nu}\,.
$$
\end{prop}
This proposition ensures that if the convex surrogate $\varphi$ is continuous, strengthening the constraint on type~I error does not deteriorate too much the optimal type~II error. We should mention that the proof does not use the Lipschitz property of $\varphi$, but only that it is uniformly bounded by $\varphi(1)$ on $[-1,1]$. This proposition has direct consequences on chance constrained programming as discussed in Section~\ref{SEC:CCP}.

The next theorem shows that the NP classifier $\tilde h^{\kappa}$ defined in subsection~\ref{sub:defin} is a good candidate to perform classification with the Neyman-Pearson paradigm. It relies on the following assumption which is necessary to verify the condition of Proposition~\ref{PROP:stochastic_constraint}.
\begin{hyp}\label{nonempty}
There exists a positive constant $\eps<1$ such that the set  of classifiers $\mathcal{H}^{\varphi, \eps\alpha}$ is nonempty.
\end{hyp}
Note that this assumption can be tested using~\eqref{EQ:inclusion_whp} for large enough $\snm$. Indeed, it follows from this inequality that with probability $1-\delta$,
$$
\mathcal{H}_{\snm}^{\varphi, \eps\alpha-\kappa/\sqrt{\snm} } \subset \mathcal{H}^{\varphi, \eps\alpha-\kappa/\sqrt{\snm} +{\kappa/\sqrt{\snm}} } = \mathcal{H}^{\varphi, \eps\alpha }\,.
$$
Thus, it is sufficient to check if $\mathcal{H}_{\snm}^{\varphi, \eps\alpha-\kappa/\sqrt{\snm} }$ is nonempty for some $\eps>0$. Before stating our main theorem, we need the following definition. Under Assumption~\ref{nonempty}, let $\bar \eps$ denote the smallest $\eps$ such that $\mathcal{H}^{\varphi, \eps\alpha} \neq \emptyset$ and let $n_0$ be the smallest integer such that
\begin{equation}
\label{EQ:n0}
n_0 \ge \left(\frac{4\kappa}{(1-\bar \eps)\alpha}\right)^2\,.
\end{equation}

\begin{TH1}
\label{TH:typeII}
Let $\varphi$, $\kappa$, $\delta$ and $\alpha$ be the same as in Theorem~\ref{TH:typeI}, and $\tilde h^{\kappa}$ denote any solution to~\eqref{EQ:NPclassifier}.  Moreover, let  Assumption~\ref{nonempty} hold and assume that  $n^{\scriptscriptstyle -}\ge n_0$ where $n_0$ is defined in~\eqref{EQ:n0}.
Then, the following hold with probability $1-2\delta$,
\begin{equation}\label{EQ:thmain1}
R^{{\scriptscriptstyle -}}(\tilde h^{\kappa})\le \Rm(\tilde h^{\kappa})\le \alpha
\end{equation}
and
\begin{equation}\label{EQ:thmain2}
\Rp(\tilde h^{\kappa})-\min_{h\in\mathcal{H}^{\varphi, \alpha}}\Rp(h)
\leq \frac{4\varphi(1)\kappa}{(1-\bar \eps)\alpha\sqrt{\snm}}+\frac{2\kappa}{\sqrt{\snp}}\,.
\end{equation}
In particular, as $M$, $\snm$ and $\snp$ all go to infinity and other quantities are held fixed,~\eqref{EQ:thmain2} yields
$$
\Rp(\tilde h^{\kappa})-\min_{h\in\mathcal{H}^{\varphi, \alpha}}\Rp(h)=\cO\left(\sqrt{\frac{\log M}{\snm}}+\sqrt{\frac{\log M}{\snp}}\right)
$$
\end{TH1}
Note here that Theorem $4.2$ is not exactly of the type \eqref{NP oracle}. The right hand side of~\eqref{EQ:thmain2} goes to zero if both $\snm$ and $\snp$ go to infinity.  Moreover, inequality~\eqref{EQ:thmain2} conveys a message that accuracy of the estimate depends on information from both classes of labeled data. This concern motivates us to consider a different sampling scheme.

\subsection{A Different Sampling Scheme}
\label{sub:sampling}

We now consider a model for observations that is more standard in statistical learning theory \citep[see, e.g.,][]{DevGyoLug96, BouBouLug05}.

Let $(X_1,Y_1), \ldots, (X_n, Y_n)$ be $n$ independent copies of the random couple $(X, Y) \in \cX \times \{-1, 1\}$. Denote by $P_X$ the marginal distribution of $X$ and by $\eta(x)=\E[Y|X=x]$ the regression function of $Y$ onto $X$. Denote by $p$ the probability of positive label and observe that
$$
p=\p[Y=1]=\E\left(\p[Y=1|X]\right)=\frac{1+\E[\eta(X)]}{2}\,.
$$
In what follows, we assume that $P_X(\eta(X)=-1)\vee P_X(\eta(X)=1) <1$ so that $p \in (0,1)$.

Let $\Nm=\card\{Y_i:Y_{i}=-1\}$ be the random number of instances labeled $-1$ and $\Np=n-\Nm=\card\{Y_i:Y_{i}=1\}$. In this setup, the NP classifier  is defined as in subsection~\ref{sub:defin} where $\snm$ and $\snp$ are replaced by $\Nm$ and $\Np$ respectively. To distinguish this classifier from $\tilde h^\tau$ previously defined, we denote the NP classifier obtained with this sampling scheme by $\tilde h_n^\tau$.

Let the event $\cF$ be defined by
$$
\cF=\{\Rm(\tilde h_n^{\kappa})\le \alpha\}\cap \{\Rp(\tilde h_n^{\kappa})-\min_{h\in\mathcal{H}^{\varphi, \alpha}}\Rp(h)
\leq\frac{4\varphi(1)\kappa}{(1-\bar \eps)\alpha\sqrt{\Nm}}+\frac{2\kappa}{\sqrt{\Np}}\}.
$$
Denote $\mathcal{B}_{\snm}=\{Y_1=\cdots=Y_{\snm}=-1, Y_{\snm+1}=\cdots=Y_n=1\}$.  Although the event $\mathcal{B}_{\snm}$ is different from the event $\{\Nm=\snm\}$, symmetry leads to the following key observation:
$$
\p(\cF|\Nm=\snm)=\p(\cF|\mathcal{B}_{\snm}).
$$


Therefore, under the conditions of Theorem~\ref{TH:typeII}, we find that for $\snm\geq n_0$ the event $\cF$
satisfies
\begin{equation}
\label{EQ:pF_cond}
\p(\cF|\Nm=\snm)\ge 1-2\delta\,.
\end{equation}
We obtain the following corollary of Theorem~\ref{TH:typeII}.
\begin{cor}
\label{COR:sampling}
Let $\varphi$, $\kappa$, $\delta$ and $\alpha$ be the same as in Theorem~\ref{TH:typeI}, and $\tilde h_n^{\kappa}$ be the NP classifier obtained with the current sampling scheme.  Then under Assumption~\ref{nonempty}, if $n > 2n_0/(1-p)$, where $n_0$ is defined in~\eqref{EQ:n0}, we have with probability $(1-2\delta)(1-e^{-\frac{n(1-p)^2}{2}})$,
\begin{equation}\label{EQ:cormain1}
R^{{\scriptscriptstyle -}}(\tilde h^{\kappa}_n)\le \Rm(\tilde h^{\kappa}_n)\le \alpha
\end{equation}
and
\begin{equation}\label{EQ:cormain2}
\Rp(\tilde h^{\kappa}_n)-\min_{h\in\mathcal{H}^{\varphi, \alpha}}\Rp(h)
\leq \frac{4\varphi(1)\kappa}{(1-\bar \eps)\alpha\sqrt{\Nm}}+\frac{2\kappa}{\sqrt{\Np}}\,.
\end{equation}
Moreover, with probability $1-2\delta -  e^{-\frac{n(1-p)^2}{2}}- e^{-\frac{np^2}{2}}$, we have simultaneously~\eqref{EQ:cormain1} and
\begin{equation}\label{EQ:cormain3}
\Rp(\tilde h^{\kappa}_n)-\min_{h\in\mathcal{H}^{\varphi, \alpha}}\Rp(h)
\leq \frac{4\sqrt{2}\varphi(1)\kappa}{(1-\bar \eps)\alpha\sqrt{n(1-p)}}+\frac{2\sqrt{2}\kappa}{\sqrt{np}}\,.
\end{equation}
\end{cor}

\section{Chance constrained optimization}
\label{SEC:CCP}
\setcounter{equation}{0}

Implementing the Neyman-Pearson paradigm for the convexified binary classification bears strong connections with chance constrained optimization. A recent account of such problems can be found in~\citet[Chapter~2]{BenEl-Nem09} and we refer to this book for references and applications. A chance constrained optimization problem is of the following form:
\begin{equation}\label{EQ:CCP}
\min_{\lambda \in \Lambda}f(\lambda) \quad  \text{s.t.} \quad \p\{F(\lambda, \xi)\leq 0\}\geq 1-\alpha,
\end{equation}
where $\xi \in \Xi$ is a random vector, $\Lambda\subset \mathbb{R}^M$ is convex, $\alpha$ is a small positive number and $f$ is a deterministic real valued convex function.  Problem~\eqref{EQ:CCP} can be viewed as a relaxation of robust optimization. Indeed, for the latter, the goal is to solve the problem
\begin{equation}
\min_{\lambda \in \Lambda}f(\lambda) \quad  \text{s.t.} \quad \sup_{\xi \in \Xi}F(\lambda, \xi)\leq 0\,,
\end{equation}
and this essentially corresponds to \eqref{EQ:CCP} for the case $\alpha=0$.
For simplicity, we take $F$ to be scalar valued but extensions to vector valued functions and conic orders are considered in ~\citet[see, e.g.,][Chapter~10]{BenEl-Nem09}. Moreover, it is standard to assume that $F(\cdot, \xi)$ is convex almost surely.

Problem~\eqref{EQ:CCP} may not be convex because the chance constraint $\{\lambda \in \Lambda\,:\, \p\{F(\lambda, \xi)\leq 0\}\geq 1-\alpha\}$ is not convex in general and thus may not be tractable. To solve this problem, \citet{Pre95} and \citet{LagLiSzn05} have derived sufficient conditions on the distribution of $\xi$ for the chance constraint to be convex. On the other hand, \citet{CalCam06} initiated a different treatment of the problem where no assumption on the distribution of $\xi$ is made, in line with the spirit of statistical learning. In that paper, they introduced the so-called \emph{scenario approach}  based on a sample $\xi_1, \ldots, \xi_n$ of independent copies of $\xi$. The scenario approach consists of solving
\begin{equation}
\label{EQ:scenario}
\min_{\lambda \in \Lambda}f(\lambda) \quad  \text{s.t.} \quad  F(\lambda,\xi_i )\leq 0, i = 1, \ldots, n.
\end{equation}
\citet{CalCam06} showed that under certain conditions, if the sample size $n$ is bigger than some $n(\alpha,\delta)$, then with probability $1-\delta$, the optimal solution $\hat \lambda^{sc}$ of \eqref{EQ:scenario} is feasible for \eqref{EQ:CCP}. The authors did not address the control of the term $f (\hat \lambda^{sc})-f^*$ where $f^*$ denotes the optimal objective value in~\eqref{EQ:CCP}. However, in view of Proposition~\ref{PROP:cannon}, it is very unlikely that this term can be controlled well.

In an attempt to overcome this limitation, a new \emph{analytical} approach  was introduced by~\citep{NemSha06}. It amounts to solving the following convex optimization problem
\begin{equation}
\label{EQ:cvxapp}
\min_{\lambda\in \Lambda, t\in\mathbb{R}^s}f(\lambda) \quad  \text{s.t.} \quad  G(\lambda, t)\leq 0,
\end{equation}
in which $t$ is some additional instrumental variable and where $G(\cdot, t)$ is convex. The problem \eqref{EQ:cvxapp} provides a conservative convex approximation to \eqref{EQ:CCP}, in the sense that  every $x$ feasible for  \eqref{EQ:cvxapp} is also feasible for \eqref{EQ:CCP}. \citet{NemSha06} considered a particular class of conservative convex approximation where the key step is to replace $\p\{F(\lambda, \xi)\ge 0\}$ by $ \E \varphi(F(\lambda, \xi))$ in~\eqref{EQ:CCP}, where $\varphi$ a nonnegative, nondecreasing, convex function that takes value $1$ at $0$.
\citet{NemSha06} discussed several choices of $\varphi$ including hinge and exponential losses, with a focus on the latter that they name \emph{Bernstein Approximation}.

The idea of a  conservative convex approximation  is also what we employ in our paper. Recall that $P^{-}$ the conditional distribution of $X$ given $Y=-1$.  In a parallel form of \eqref{EQ:CCP}, we cast our target problem as
\begin{equation}
\label{EQ:NPCCP}
\min_{\lambda\in\Lambda}R^{{\scriptscriptstyle +}}(\hh_\lambda)  \quad  \text{s.t.} \quad \Pm\{\hh_\lambda(X)\leq0\}\geq 1-\alpha,
\end{equation}
where $\Lambda$ is the flat simplex of $\R^M$.

Problem \eqref{EQ:NPCCP} differs from \eqref{EQ:CCP} in that $R^{{\scriptscriptstyle +}}(\hh_\lambda)$ is not a convex function of $\lambda$.  Replacing $R^{{\scriptscriptstyle +}}(\hh_\lambda)$ by $\Rp(\hh_\lambda)$ turns \eqref{EQ:NPCCP} into a standard chance constrained optimization problem:
\begin{equation}
\label{EQ:std}
\min_{\lambda\in\Lambda}\Rp(\hh_\lambda)  \quad  \text{s.t.} \quad \Pm\{\hh_\lambda(X)\leq0\}\geq 1-\alpha.
\end{equation}
However, there are two important differences in our setting, so that we cannot use directly Scenario Approach or  Bernstein Approximation or other analytical approaches to \eqref{EQ:CCP}. First, $\Rp(f_\lambda)$ is an \emph{unknown} function of $\lambda$.  Second, we assume minimum knowledge about $\Pm$.  On the other hand, chance constrained optimization techniques in previous literature assume knowledge about the distribution of the random vector $\xi$. For example, \citet{NemSha06} require that  the moment generating function of the random vector $\xi$ is efficiently computable to study the Bernstein Approximation.

Given a finite sample, it is not feasible to construct a strictly conservative approximation to the constraint in \eqref{EQ:std}. Instead, what possible is to ensure that if we learned $\hat{\hh}$ from the sample, this constraint is satisfied with high probability $1-\delta$, i.e., the classifier is approximately feasible for \eqref{EQ:std}. In retrospect, our approach to \eqref{EQ:std} is an innovative hybrid between the analytical approach based on convex surrogates and the scenario approach.

We do have structural assumptions on the problem.  Let $g_j, j\in\{1,\ldots, M\}$ be arbitrary functions that take values in $[-1,1]$  and $F(\lambda,\xi)=\sum_{j=1}^N \lambda_j g_j(\xi)$. Consider a convexified version of \eqref{EQ:CCP}:
\begin{equation}
\label{EQ:cvx_CCP}
\min_{\lambda \in\Lambda}f(\lambda)  \quad  \text{s.t.} \quad  \E[\varphi(F(\lambda,\xi))]\leq\alpha,
\end{equation}
where $\varphi$ is a $L$-Lipschitz convex surrogate, $L>0$. Suppose that we observe a sample $(\xi_1, \ldots, \xi_n)$ that are independent copies of $\xi$. We propose to approximately solve the above problem by
$$
\min_{\lambda \in\Lambda}f(\lambda)  \quad  \text{s.t.} \quad  \sum_{i=1}^n\varphi(F(\lambda,\xi_i))\leq n\alpha-\kappa\sqrt{n}\,,
$$
for some $\kappa>$ to be defined. Denote by $\tilde \lambda$ any solution to this problem  and by $f^*_\varphi$ the value of the objective at the optimum in~\eqref{EQ:cvx_CCP}.
The following theorem summarizes our contribution to chance constrained optimization.
\begin{TH1}
\label{TH:CCP}
Fix constants $\delta, \alpha\in(0,1/2), L>0$ and let $\varphi:[-1,1]\to \R^+$ be a given $L$-Lipschitz convex surrogate. Define
$$
\kappa=4\sqrt{2}L\sqrt{\log \left(\frac{2M}{\delta}\right)}\,.
$$
Then, the following hold with probability at least $1-2\delta$
\begin{itemize}
\item[(i)] $\tilde \lambda$ is feasible for~\eqref{EQ:CCP}.
\item[(ii)] If there exists $\eps \in (0,1)$ such that the constraint $\E[\varphi(F(\lambda,\xi))]\leq\eps\alpha$ is feasible for some $\lambda \in \Lambda$, then for
$$
n \ge \left(\frac{4\kappa}{(1-\eps)\alpha}\right)^2\,,
$$
we have
$$
f(\tilde \lambda)-f^*_\varphi \le \frac{4\varphi(1)\kappa}{(1- \eps)\alpha\sqrt{n}}\,.
$$
In particular, as $M$ and $n$ go to infinity with all other quantities kept fixed, we obtain
$$
f(\tilde \lambda)-f^*_\varphi=\cO\left(\sqrt{\frac{\log M}{n}}\right)\,.
$$
\end{itemize}
\end{TH1}
The proof essentially follows that of Theorem~\ref{TH:typeII} and we omit it.
The limitations of Theorem~\ref{TH:CCP} include rigid structural assumptions on the function $F$ and on the set $\Lambda$. While the latter can be easily relaxed using more sophisticated empirical process theory, the former is inherent to our analysis. Also, we did not address the effect of replacing the indicator function by a convex surrogate; this investigation is beyond the scope of this paper.
\section{Appendix}
\setcounter{equation}{0}

\subsection{Proof of Proposition~\ref{PROP:cannon}}
Let the base classifiers be defined as
$$
h_1(x)= -1 \quad \text{and}\quad h_2(x)=\1(x\leq\alpha)-\1(x>\alpha)\,,\quad \forall\, x \in [0,1]
$$
For any $\lambda \in [0,1]$, denote the convex combination of $h_1$ and $h_2$ by $\hh_\lambda=\lambda h_1 +(1-\lambda)h_2$, i.e.,
$$
\hh_\lambda(x)=(1-2\lambda)\1(x\le \alpha) -\1(x>\alpha)\,.
$$
Suppose the conditional distributions of $X$ given $Y=1$ or $Y=-1$, denoted respectively by $\Pp$ and $\Pm$, are both uniform on $[0,1]$. Recall that
${R}^{{\scriptscriptstyle -}}(\hh_\lambda)={P}^{{\scriptscriptstyle -}}(\hh_\lambda(X) \ge 0)$ and ${R}^{{\scriptscriptstyle +}}(\hh_\lambda)={P}^{{\scriptscriptstyle +}}(\hh_\lambda(X) \le 0)$\,.
Then, we have
\begin{equation}
\label{EQ:pr_p31_1}
{R}^{{\scriptscriptstyle -}}(\hh_\lambda)=\Pm(\hh_\lambda(X) \ge 0)=\alpha \1(\lambda\le 1/2)\,.
\end{equation}

Therefore, for any $\tau \in [0, \alpha]$, we have
$$
\{\lambda \in [0,1]\,:\, {R}^{{\scriptscriptstyle -}}(\hh_\lambda)\le \tau\}=
\left\{
\begin{array}{ll}
\left[0,1\right] & \text{if}\ \tau=\alpha\,,\\
\left(1/2,1\right]&  \text{if}\ \tau<\alpha\,.\\
\end{array}\right.
$$
Observe now that
\begin{equation}
\label{EQ:pr_p31_2}
{R}^{{\scriptscriptstyle +}}(\hh_\lambda)=\Pp(\hh_\lambda(X) \le 0)=(1-\alpha) \1(\lambda< 1/2)+ \1(\lambda\ge 1/2)\,.
\end{equation}
For any $\tau \in [0, \alpha]$, it yields
$$
\inf_{\lambda \in [0,1]:{R}^{{\scriptscriptstyle -}}(\hh_\lambda)\leq\tau}{R}^{{\scriptscriptstyle +}}(\hh_\lambda)=\left\{
\begin{array}{ll}
1-\alpha & \text{if}\ \tau=\alpha\,,\\
1 &  \text{if}\ \tau<\alpha\,.\\
\end{array}\right.
$$
Consider now a classifier $\bar{\hh}_\lambda$ such that ${R}^{{\scriptscriptstyle -}}(\bar{\hh}_\lambda)\leq\tau$ for some $\tau <\alpha$. Then from~\eqref{EQ:pr_p31_1}, we see that must have $\lambda>1/2$. Together with~\eqref{EQ:pr_p31_2}, this imples that ${R}^{{\scriptscriptstyle +}}(\bar{\hh}_\lambda)=1$. It yields
$$
{R}^{{\scriptscriptstyle +}}(\bar{\hh}_\lambda) - \min_{\lambda \,: \, {R}^{{\scriptscriptstyle -}}(\hh_\lambda)\le \alpha}{R}^{{\scriptscriptstyle +}}(\hh_\lambda)=1-(1-\alpha)=\alpha\,.
$$
This completes the first part of the proposition.
Moreover, in the same manner as~\eqref{EQ:pr_p31_1}, it can be easily proved that
\begin{equation}
\label{EQ:pr_p31_1_2}
{\hat R}^{{\scriptscriptstyle -}}(\hh_\lambda)=\frac{1}{\snm}\sum_{i=1}^{\snm}\1(\hh_\lambda(X_i^{{\scriptscriptstyle -}}) \ge 0)=\alpha_{\snm} \1(\lambda\le 1/2)\,,
\end{equation}
where
\begin{equation}
\label{EQ:alpha_n}
\alpha_{\snm} =\frac{1}{\snm}\sum_{i=1}^{\snm}\1(X_i^{{\scriptscriptstyle -}}\le \alpha)
\end{equation}
If a classifier $\hat \hh_{\lambda}$ is such that  ${\hat R}^{{\scriptscriptstyle -}}(\hat {\hh}_\lambda) <\alpha_{\snm}$, then~\eqref{EQ:pr_p31_1_2} implies that $\lambda > 1/2$. Using again~\eqref{EQ:pr_p31_2}, we find also that ${R}^{{\scriptscriptstyle +}}({\hat \hh}_\lambda)=1$.  It yields
$$
{R}^{{\scriptscriptstyle +}}(\hat{\hh}_\lambda) - \min_{\lambda \,: \, {R}^{{\scriptscriptstyle -}}(\hh_\lambda)\le \alpha}{R}^{{\scriptscriptstyle +}}(\hh_\lambda)=1-(1-\alpha)=\alpha\,.
$$
It remains to show that ${\hat R}^{{\scriptscriptstyle -}}(\hat {\hh}_\lambda) <\alpha_{\snm}$ with positive probability for any classifier such that ${\hat R}^{{\scriptscriptstyle -}}(\hat {\hh}_\lambda)\leq\tau$ for some $\tau <\alpha$.  Note that a sufficient condition for a classifier $\hat {\hh}_\lambda$ to satisfy this constraint is to have  $\alpha\le \alpha_{\snm}$. It is therefore sufficient to find a lower bound on the probability of the event $\cA=\{\alpha_{\snm} \ge \alpha\}$. Such a lower bound is provided by Lemma~\ref{LEM:bin2}, which guarantees that $\p(\cA)\ge \alpha\wedge 1/4$.

\subsection{Proof of Theorem~\ref{TH:typeI}}
We begin with the following lemma, which is extensively used in the sequel. Its proof relies on standard arguments to bound suprema of empirical processes. Recall that $\{h_1, \ldots, h_M\}$ is family of $M$ classifiers such that $h_j:\cX \to [-1, 1]$  and  that for any $\lambda$ in the simplex $\Lambda \subset R^M$, $\hh_\lambda$ denotes the convex combination defined by
$$
\hh_\lambda = \sum_{j=1}^N \lambda_j h_j\,.
$$
The following standard notation in empirical process theory will be used.  Let $X_1, \ldots, X_n\in\cX$ be $n$ i.i.d random variables with marginal distribution $P$. Then for any measurable function $f:\cX \to \R$, we write
$$
P_n (f)=\frac{1}{n}\sum_{i=1}^n f(X_i)\qquad \text{and} \qquad P(f)=\E f(X)=\int f \ud P\,.
$$
Moreover, the Rademacher average of $f$ is defined as
$$
R_n(f)=\frac{1}{n}\sum_{i=1}^n \varepsilon_i f(X_i)\,,
$$
where $\varepsilon_1, \ldots, \varepsilon_n$ are i.i.d. Rademacher random variables such that $\p(\varepsilon_i=1)=\p(\varepsilon_i=-1)=1/2$ for $i=1, \ldots, n$.
\begin{lem}
\label{LEM:sup}
Fix $L>0, \delta \in (0,1)$. Let $X_1, \ldots, X_n$ be $n$ i.i.d random variables on $\cX$ with marginal distribution $P$. Moreover, let $\varphi: [-1,1] \to \R$ an $L$-Lipschitz function. Then, with probability at least $1-\delta$, it holds
$$
\sup_{\lambda \in \Lambda}\left|(P_n -P) (\varphi \circ \hh_\lambda)  \right| \le \frac{4\sqrt{2}L}{\sqrt{n}}\sqrt{\log \left(\frac{2M}{\delta}\right)}\,.
$$
\end{lem}
{\sc Proof.}
Define $\bar \varphi (\cdot) \doteq \varphi(\cdot)-\varphi(0)$, so that $\bar \varphi$ is an $L$-Lipschitz function that satisfies $\bar \varphi(0)=0$. Moreover, for any $\lambda \in \Lambda$, it holds
$$
(P_n -P)( \varphi \circ \hh_\lambda) =(P_n -P) (\bar \varphi \circ \hh_\lambda )\,.
$$
Let $\Phi: \R \to \R_+$ be a given convex increasing function. Applying successively the symmetrization and the contraction inequalities \citep[see, e.g.,][Section~2]{Kol08}, we find
$$
\E\Phi\left( \sup_{\lambda \in \Lambda}\left|(P_n -P) (\bar \varphi \circ \hh_\lambda)  \right|\right)\le \E\Phi\left(2\sup_{\lambda \in \Lambda}\left|R_n( \bar \varphi \circ \hh_\lambda)  \right|\right)\le  \E\Phi\left(4L\sup_{\lambda \in \Lambda}\left|R_n( \hh_\lambda)  \right|\right)\,.
$$
Observe now that $\lambda \mapsto \left|R_n( \hh_\lambda)  \right|$ is a convex function and Theorem~32.2 in \citet{Roc97} entails that
$$
\sup_{\lambda \in \Lambda}\left|R_n( \hh_\lambda)  \right|=\max_{1\le j \le M} \left|R_n( h_j)  \right|\,.
$$
We now use a Chernoff bound to control this quantity. To that end, fix $s,t>0$, and observe that
\begin{align}
\p\left(\sup_{\lambda \in \Lambda}\left|(P_n -P) (\varphi \circ \hh_\lambda)  \right|>t \right)& \le\frac{1}{\Phi(st) }\E\Phi\left( s\sup_{\lambda \in \Lambda}\left|(P_n -P) (\bar \varphi \circ \hh_\lambda)  \right|\right) \nonumber&\\
& \le \frac{1}{\Phi(st) }\E\Phi\left(4Ls\max_{1\le j \le M} \left|R_n( h_j)  \right|\right) \,. \label{EQ:pr_lemsup_1}&
\end{align}
Moreover, since $\Phi$ is increasing,
\begin{align}
\E\Phi\left(4Ls\max_{1\le j \le M} \left|R_n( h_j)  \right|\right) &=\E\max_{1\le j \le M}\Phi\left(4Ls \left|R_n( h_j)  \right|\right)\nonumber&\\
&\le \sum_{j=1}^M\E \left[\Phi\left(4Ls R_n( h_j) \right)\vee \Phi\left(-4Ls R_n( h_j) \right)\right]\nonumber&\\
&\le 2\sum_{j=1}^M\E \Phi\left(4Ls R_n( h_j) \right)\label{EQ:pr_lemsup_2}\,.
\end{align}
Now choose $\Phi(\cdot)=\exp(\cdot)$, then
$$
\E \Phi\left(4Ls R_n( h_j) \right)=	\prod_{i=1}^n \E  \cosh\left(\frac{4Lsh_j(X_i)}{n}\right)\le \exp\left(\frac{8L^2 s^2}{n}\right)\,,
$$
where $\cosh$ is the hyperbolic cosine function and where in the inequality, we used the fact that $|h_j(X_i)|\le 1$ for any $i,j$ and $\cosh(x)\le \exp(x^2/2)$. Together with~\eqref{EQ:pr_lemsup_1} and~\eqref{EQ:pr_lemsup_2}, it yields
\begin{align*}
\p\left(\sup_{\lambda \in \Lambda}\left|(P_n -P) (\varphi \circ \hh_\lambda)  \right|>t \right)&\le 2M \inf_{s>0}\exp\left(\frac{8L^2 s^2}{n}-st\right)\le 2M\exp\left(-\frac{nt^2}{32L^2}\right)\,.
\end{align*}
Choosing
$$
t=\frac{4\sqrt{2}L}{\sqrt{n}}\sqrt{\log \left(\frac{2M}{\delta}\right)}\,,
$$
completes the proof of the Lemma.
\epr

We now proceed to the proof of Theorem~\ref{TH:typeI}.
Note first that from the properties of $\varphi$, $R^{{\scriptscriptstyle -}}(h) \le \Rm (h)$. Next, we have for any data-dependent classifier $h \in \HC$ such that $\hRm(h)\le \alpha_\kappa$:
$$
 \Rm( h)\le \hRm(h) + \sup_{h \in \HC}\left| \hRm(h) -  \Rm( h)\right| \le \alpha - \frac{\kappa}{\sqrt{\snm}} + \sup_{h \in \HC}\left| \hRm(h) -  \Rm( h)\right|\,.
$$
Lemma~\ref{LEM:sup} implies that, with probability $1-\delta$
$$
\sup_{h \in \HC}\left| \hRm(h) -  \Rm( h)\right|=\sup_{\lambda \in \Lambda}\left|(\Pm_{\snm} -\Pm) (\varphi \circ \hh_\lambda)  \right| \le \frac{\kappa}{\sqrt{\snm}}\,.
$$
The previous two displays imply that  $\Rm( h)\le \alpha$ with probability $1-\delta$, which completes the proof of Theorem~\ref{TH:typeI}.

\subsection{Proof of Proposition~\ref{PROP:stochastic_constraint}}
The proof of this proposition builds upon the following lemma.
\begin{lem}
\label{LEM:gamma}
Let $\gamma(\alpha)=\inf_{\hh_{\lambda}\in\mathcal{H}^{\varphi, \alpha}}\Rp(\hh_{\lambda})$, then $\gamma$ is a non-increasing convex function on $[0,1]$.
\end{lem}
{\sc Proof.}
First, it is clear that $\gamma$ is a non-increasing function of $\alpha$ because for $\alpha'>\alpha$, $\{\hh_\lambda \in \HC\,:\,\Rm(\hh_{\lambda})\leq\alpha\}\subset\{\hh_\lambda \in \HC\,:\,\Rm(\hh_{\lambda})\leq\alpha'\}$.

We now show that $\gamma$ is convex. To that end, observe first that since $\varphi$ is continuous on $[-1,1]$, the set $\{\lambda \in \Lambda\, : \, \hh_{\lambda}\in\mathcal{H}^{\varphi, \alpha}\}$ is compact. Moreover, the function $\lambda \mapsto \Rp(\hh_{\lambda})$ is convex. Therefore, there exists $\lambda^* \in \Lambda$ such that
$$
\gamma(\alpha)=\inf_{\hh_{\lambda}\in\mathcal{H}^{\varphi, \alpha}}\Rp(\hh_{\lambda})=\min_{\hh_{\lambda}\in\mathcal{H}^{\varphi, \alpha}}\Rp(\hh_{\lambda})=\Rp(\hh_{\lambda^*})\,.
$$
Now, fix $\alpha_1, \alpha_2 \in [0,1]$. From the above considerations, there exist $\lambda_1, \lambda_2 \in \Lambda$ such that $\gamma(\alpha_1)=\Rp(\hh_{\lambda_1})$ and $\gamma(\alpha_2)=\Rp(\hh_{\lambda_2})$. For any $\theta \in (0,1)$, define the convex combinations $\bar \alpha_\theta=\theta \alpha_1 + (1-\theta)\alpha_2$ and $\bar \lambda_\theta=\theta \lambda_1 + (1-\theta)\lambda_2$. Since $\lambda \mapsto \Rm(\hh_\lambda)$ is convex, it holds
$$
\Rm(\hh_{\bar \lambda_\theta})\le \theta \Rm(\hh_{\lambda_1})+(1-\theta)\Rm(\hh_{\lambda_2})\le \theta \alpha_1+(1-\theta)\alpha_2=\bar \alpha_\theta\,,
$$
so that $\hh_{\bar \lambda_\theta} \in \mathcal{H}^{\varphi, \bar \alpha_\theta}$. Hence, $\gamma( \bar \alpha_\theta)\le \Rp(\hh_{\bar \lambda_\theta})$. Together with the convexity of $\varphi$, it yields
$$
\gamma(\theta \alpha_1 + (1-\theta)\alpha_2) \le \Rp(\hh_{\bar \lambda_\theta})\le  \theta\Rp(\hh_{\lambda_1})+(1-\theta)\Rp(\hh_{\lambda_2})=\theta \gamma(\alpha_1) + (1-\theta)\gamma(\alpha_2) \,.
$$
\epr

We now complete the proof of Proposition~\ref{PROP:stochastic_constraint}.
For any $x \in [0,1]$, let $\gamma(x)=\inf_{h\in\mathcal{H}^{\varphi, x}}\Rp(h)$ and observe that the statement of the proposition is equivalent to
\begin{equation}
\label{EQ:ineq_gamma}
\gamma(\alpha - \nu)- \gamma(\alpha) \le \varphi(1)\frac{\nu}{\nu_0-\nu}\,, \quad 0<\nu <\nu_0\,.
\end{equation}
Lemma~\ref{LEM:gamma} together with the assumption that $\mathcal{H}^{\varphi, \alpha-\nu_0} \neq \emptyset$ imply that $\gamma$ is a non-increasing convex real-valued function on $[\alpha - \nu_0, 1]$ so that
$$
\gamma(\alpha - \nu)- \gamma(\alpha) \le \nu \sup_{g \in \partial \gamma(\alpha -\nu)}|g|\,,
$$
where $\partial \gamma(\alpha -\nu)$ denotes the sub-differential of $\gamma$ at $\alpha -\nu$. Moreover, since $\gamma$ is a non-increasing convex function on $[\alpha- \nu_0, \alpha-\nu]$, it holds
$$
\gamma(\alpha - \nu_0)- \gamma(\alpha-\nu)\ge (\nu -\nu_0)\sup_{g \in \partial \gamma(\alpha -\nu)}|g|\,.
$$
The previous two displays yield
$$
\gamma(\alpha - \nu)- \gamma(\alpha) \le \nu \frac{\gamma(\alpha - \nu_0)- \gamma(\alpha-\nu)}{\nu -\nu_0} \le \nu \frac{\varphi(1)}{\nu -\nu_0}\,.
$$

\subsection{Proof of Theorem~\ref{TH:typeII}}
Define the events $\cE^{{\scriptscriptstyle -}}$ and $\cE^{{\scriptscriptstyle +}}$ by
\begin{align*}
\cE^{{\scriptscriptstyle -}}&=\bigcap_{h \in \HC} \{|\hRm(h)-\Rm(h)|\le \frac{\kappa}{\sqrt{\snm}}\}\,,&\\
\cE^{{\scriptscriptstyle +}}&=\bigcap_{h \in \HC} \{|\hRp(h)-\Rp(h)|\le \frac{\kappa}{\sqrt{\snp}}\}\,.&
\end{align*}
Lemma~\ref{LEM:sup} implies
\begin{equation}
\label{EQ:probE}
\p(\cE^{{\scriptscriptstyle -}}) \wedge \p(\cE^{{\scriptscriptstyle +}})\ge 1- \delta\,.
\end{equation}
Note first that Theorem~\ref{TH:typeI} implies that~\eqref{EQ:thmain1} holds with probability $1-\delta$.
Observe now that the l.h.s of~\eqref{EQ:thmain2} can be decomposed as
$$
\Rp(\tilde h^{\kappa})-\min_{h\in\mathcal{H}^{\varphi, \alpha}}\Rp(h)= A_1 + A_1 + A_3\,,
$$
where
\begin{align*}
A_1&=\left(\Rp(\tilde h^{\kappa})- \hRp(\tilde h^{\kappa})\right) + \left(\hRp(\tilde h^{\kappa})- \min_{h\in\mathcal{H}_{\snm}^{\varphi, \alpha_\kappa}}\Rp(h)\right)&\\
A_2&=\min_{h\in\mathcal{H}_{\snm}^{\varphi, \alpha_\kappa}}\Rp(h)-\min_{h\in\mathcal{H}^{\varphi, \alpha_{2\kappa}}}\Rp(h)&\\
A_3&=\min_{h\in\mathcal{H}^{\varphi, \alpha_{2\kappa}}}\Rp(h)-\min_{h\in\mathcal{H}^{\varphi, \alpha}}\Rp(h).
\end{align*}

To bound $A_1$ from above, observe that
$$
A_1\le 2\sup_{h\in\mathcal{H}_{\snm}^{\varphi, \alpha_\kappa}} |\hRp(h)-\Rp(h)|\le 2\sup_{h\in\HC} |\hRp(h)-\Rp(h)|.
$$
Therefore, on the event $\cE^{{\scriptscriptstyle +}}$ it holds
$$
A_1\le \frac{2\kappa}{\sqrt{\snp}}\,.
$$

We now treat $A_2$. Note that $A_2\le 0$  on the event $\mathcal{H}^{\varphi, \alpha_{2\kappa}} \subset \mathcal{H}_{\snm}^{\varphi, \alpha_\kappa}$. But this event contains $\cE^{{\scriptscriptstyle -}}$ so that $A_2 \le 0$ on the event $\cE^{{\scriptscriptstyle -}}$.

Finally, to control $A_3$, observe that under Assumption~\ref{nonempty}, Proposition~\ref{PROP:stochastic_constraint} can be applied with $\nu=2\kappa/\sqrt{\snm}$ and $\nu_0=(1-\bar \eps) \alpha$. Indeed, the assumptions of the theorem imply that $\nu\le \nu_0/2$.  It yields
$$
A_3\le \frac{4\varphi(1)\kappa}{(1-\bar \eps)\alpha\sqrt{\snm}}\,.
$$
Combining the bounds on $A_1$, $A_2$ and $A_3$ obtained above, we find that~\eqref{EQ:thmain2} holds on the event $\cE^{{\scriptscriptstyle -}}\cap \cE^{{\scriptscriptstyle +}}$ that has probability at least $1-2\delta$ in view of~\eqref{EQ:probE}.

The last statement of the theorem follows directly from the definition of $\kappa$.

\subsection{Proof of Corollary~\ref{COR:sampling}}
Now prove~\eqref{EQ:cormain2},
\begin{align*}
\p(\cF)&=\sum_{\snm=0}^n \p(\cF|\Nm=\snm)\p(\Nm=\snm) &\\
&\ge \sum_{\snm=n_0}^n \p(\cF|\Nm=\snm)\p(\Nm=\snm) &\\
&\ge (1-2\delta) \p(\Nm\ge n_0)\,,
\end{align*}
where in the last inequality, we used~\eqref{EQ:pF_cond}. Applying now Lemma~\ref{LEM:bin}, we obtain
$$
\p(\Nm\ge n_0)\ge 1-e^{-\frac{n(1-p)^2}{2}}\,.
$$
Therefore,
$$
\p(\cF)\ge (1-2\delta)(1-e^{-\frac{n(1-p)^2}{2}})\,,
$$
which completes the proof of~\eqref{EQ:cormain2}.

The proof of~\eqref{EQ:cormain3} follows by observing that
$$
\left\{\Rp(\tilde h^{\kappa}_n)-\min_{h\in\mathcal{H}^{\varphi, \alpha}}\Rp(h)
> \frac{4\sqrt{2}\varphi(1)\kappa}{(1-\bar \eps)\alpha\sqrt{n(1-p)}}+\frac{2\sqrt{2}\kappa}{\sqrt{np}}\right\} \subset \cA_1 \cup \cA_2 \cup \cA_3=(\cA_1 \cap \cA_2^c)\cup \cA_2 \cup \cA_3\,,
$$
where
\begin{align*}
\cA_1&=\left\{\Rp(\tilde h^{\kappa}_n)-\min_{h\in\mathcal{H}^{\varphi, \alpha}}\Rp(h)
> \frac{4\varphi(1)\kappa}{(1-\bar \eps)\alpha\sqrt{\Nm}}+\frac{2\kappa}{\sqrt{\Np}}\right\} \subset \cF^c\,,&\\
\cA_2&=\{ \Nm  < n(1-p)/2\}   \,,&\\
\cA_3&=  \{ \Np <  np/2\} \,.&
\end{align*}
Since $\cA_2^c \subset \{\Nm \ge n_0\}$, we find
$$
\p(\cA_1 \cap \cA_2^c)\le \sum_{\snm\ge n_0}\p(\cF^c|\Nm=\snm)\p(\Nm=\snm)\le 2\delta\,.
$$
Next, using Lemma~\ref{LEM:bin}, we get
$$
\p(\cA_2)\le e^{-\frac{n(1-p)^2}{2}}\qquad \text{and} \qquad \p(\cA_3)\le e^{-\frac{np^2}{2}}\,.
$$
Hence, we find
$$
\p\left\{\Rp(\tilde h^{\kappa}_n)-\min_{h\in\mathcal{H}^{\varphi, \alpha}}\Rp(h)
> \frac{4\sqrt{2}\varphi(1)\kappa}{(1-\bar \eps)\alpha\sqrt{n(1-p)}}+\frac{2\sqrt{2}\kappa}{\sqrt{np}}\right\}\le 2\delta +  e^{-\frac{n(1-p)^2}{2}}+e^{-\frac{np^2}{2}}\,,
$$
which completes the proof of the corollary.

\subsection{Technical lemmas on Binomial distributions}

The following lemmas are purely technical and arise from the fact that we observe binary data. They are used in two unrelated results.
\begin{lem}
\label{LEM:bin}
Let $N$ be a binomial random variables with parameters $n \ge 1$ and $q \in (0,1)$. Then, for any $t>0$ such that $t\le nq/2$, it holds
$$
\p(N \ge t) \ge 1-e^{-\frac{nq^2}{2}}\,.
$$
\end{lem}
{\sc Proof.} Note first that $n-N$ has binomial distribution with parameters $n \ge 1$ and $1-q$. Therefore, we can write $n-N=\sum_{i=1}^n Z_i$ where $Z_i$ are i.i.d. Bernoulli random variables with parameter $1-q$. Thus, using Hoeffding's inequality, we find that for any $s\ge 0$,
$$
\p(n-N - n(1-q)\ge s)\le e^{-\frac{2s^2}{n}}\,.
$$
Applying the above inequality with $s=n-n(1-q)-t\ge nq/2\ge 0$ yields
$$
\p(N\ge t)=\p(n-N - n(1-q)\le n -n(1-q) - t)\ge 1-e^{-\frac{nq^2}{2}}\,.
$$
\epr
The next lemma provides a lower bound on the probability that a binomial distribution exceeds its expectation. Our result is uniform in the size of the binomial and it can be easily verified that it is sharp by considering sizes $n=1$ and $n=2$. In particular, we do resort to Gaussian approximation which improves upon the lower bounds that can be derived from the inequalities presented in~\citet{Slu77}.
\begin{lem}
\label{LEM:bin2}
Let $N$ be a binomial random variable with parameters $n \ge 1$ and $0<q\le 1/2$. Then, it holds
$$
\p(N \ge nq) \ge
q\wedge  (1/4)\,.$$
\end{lem}
{\sc Proof.} We introduce the following local definition, which is limited to the scope of the this proof. Fix $n \ge 1$ and for any $q \in (0,1)$, let $P_{q}$ denote the distribution of a binomial random variable with parameters $n$ and $q$. Note first that if $n=1$, the result is trivial since
$$
P_{ q}(N \ge q)=\p(Z \ge q)=\p(Z=1)=q\,,
$$
where $Z$ is a Bernoulli random variable with parameter $q$.

Assume that $n \ge 2$.
Note that if $q\le 1/n$, then $P_q(N\ge nq)\ge \p(Z=1)=q$, where $Z$ is a Bernoulli random variable with parameter $q$. Moreover, for
any any integer $k$ such that $k/n< q \le (k+1)/n$, we have
\begin{equation}
\label{EQ:pr_bin2_0}
P_{q}(N \ge nq) = P_{q}(N \ge k+1)\ge P_{\frac{k}{n}}(N \ge  k+1)\,.
\end{equation}
The above inequality can be easily proved by taking the derivative over the interval $(k/n,(k+1)/n]$, of the function
$$
q \mapsto \sum_{j=k+1}^n{n \choose j}q^j(1-q)^j\,.
$$
We now show that
\begin{equation}
\label{EQ:pr_bin2_1}
P_{\frac{k}{n}}(N \ge  k+1) \ge P_{\frac{k-1}{n}}(N \ge  k)\,,\quad 2\le k \le n/2\,.
\end{equation}

Let $U_1, \ldots, U_n$ be $n$ i.i.d. random variables uniformly distributed on the interval $[0,1]$ and denote by $U_{(k)}$ the corresponding $k$th order statistic such that $U_{(1)} \le \ldots \le U_{(n)}$. Following~\citet[Section~7.2]{Fel71}, it is not hard to show that
$$
P_{\frac{k}{n}}(N \ge  k+1) =\p(U_{(k+1)}\le \frac{k}{n})=n{n-1 \choose k}\int_0^{\frac{k}{n}}t^k (1-t)^{n-k-1}\ud t\,,
$$
and in the same manner,
$$
P_{\frac{k-1}{n}}(N \ge  k) =\p(U_{(k)}\le \frac{k-1}{n})=n{n-1 \choose k-1}\int_0^{\frac{k-1}{n}}t^{k-1} (1-t)^{n-k}\ud t\,.
$$
Note that
$$
{n-1 \choose k-1}={n-1 \choose k}\frac{k}{n-k}\,,
$$
so that~\eqref{EQ:pr_bin2_1} follows if we prove
\begin{equation}
\label{EQ:pr_bin2_2}
k\int_0^{\frac{k-1}{n}}t^{k-1} (1-t)^{n-k}\ud t \le (n-k)\int_0^{\frac{k}{n}}t^k (1-t)^{n-k-1}\ud t\,.
\end{equation}
We can establish the following chain of equivalent inequalities.
\begin{align*}
&&k\int_0^{\frac{k-1}{n}}t^{k-1} (1-t)^{n-k}\ud t &\le (n-k)\int_0^{\frac{k}{n}}t^k (1-t)^{n-k-1}\ud t&\\
&\Leftrightarrow & \int_0^{\frac{k}{n}} \frac{\ud t^k}{\ud t}(1-t)^{n-k}\ud t   &\le -\int_0^{\frac{k}{n}}t^k \frac{\ud (1-t)^{n-k}}{\ud t}\ud t+ k\int_{\frac{k-1}{n}}^{\frac{k}{n}}t^{k-1} (1-t)^{n-k}\ud t&\\
&\Leftrightarrow & \int_0^{\frac{k}{n}} \frac{\ud }{\ud t}\left[t^k(1-t)^{n-k}\right]\ud t  &\le k\int_{\frac{k-1}{n}}^{\frac{k}{n}}t^{k-1} (1-t)^{n-k}\ud t&\\
&\Leftrightarrow & \left(\frac{k}{n}\right)^{k} \left(1-\frac{k}{n}\right)^{n-k}  &\le k\int_{\frac{k-1}{n}}^{\frac{k}{n}}t^{k-1} (1-t)^{n-k}\ud t&
\end{align*}
We now study the variations of the function $t\mapsto b(t)=t^{k-1} (1-t)^{n-k}$ on the interval $[(k-1)/n, k/n]$. Taking derivative, it is not hard to see that function $b$ admits a unique local optimum, which is a maximum, at $t_0=\frac{k-1}{n-1}$ and that $t_0  \in ((k-1)/n, k/n)$ because $k \le n$. Therefore, the function is increasing on $[(k-1)/n, t_0]$ and decreasing on $[t_0, k/n]$. It implies that
$$
\int_{\frac{k-1}{n}}^{\frac{k}{n}}b(t)\ud t \ge \frac{1}{n}\min\left[b\big(\frac{k-1}{n}\big), b\big(\frac{k}{n}\big) \right]\,.
$$
Hence, the proof of~\eqref{EQ:pr_bin2_2} follows from the following two observations:
$$
\left(\frac{k}{n}\right)^{k} \left(1-\frac{k}{n}\right)^{n-k} = \frac{k}{n}\left(\frac{k}{n}\right)^{k-1} \left(1-\frac{k}{n}\right)^{n-k}=\frac{k}{n}b\big(\frac{k}{n}\big)\,,
$$
and
$$
\left(\frac{k}{n}\right)^{k} \left(1-\frac{k}{n}\right)^{n-k} \le  \frac{k}{n}\left(\frac{k-1}{n}\right)^{k-1} \left(1-\frac{k-1}{n}\right)^{n-k}=\frac{k}{n}b\big(\frac{k-1}{n}\big)\,.
$$
While the first equality above is obvious, the second inequality can be obtained by an equivalent statement is\begin{align*}
&& \left(\frac{k}{n}\right)^{k-1} \left(\frac{n-k}{n}\right)^{n-k}  &\le \left(\frac{k-1}{n}\right)^{k-1} \left(\frac{n-k+1}{n}\right)^{n-k}&\\
&\Leftrightarrow & \left(\frac{k}{k-1}\right)^{k-1} \left(\frac{n-k}{n-k+1}\right)^{n-k}  &\le 1&
\end{align*}
Since the function $t\mapsto \left(\frac{t+1}{t}\right)^{t}$ is increasing on $[0,\infty)$, and $k\le n-k+1$, the result follows.

To conclude the proof of the Lemma, note that~\eqref{EQ:pr_bin2_0} and~\eqref{EQ:pr_bin2_1} imply that for any $q>1/n$,
$$
P_{q}(N \ge nq) \ge P_{\frac{1}{n}}(N \ge 2)=1-\left(\frac{n-1}{n}\right)^n-\left(\frac{n-1}{n}\right)^{n-1}\ge 1-\left(\frac{1}{2}\right)^2-\frac{1}{2}=\frac{1}{4}\,,
$$
where, in the last inequality, we used the fact that the function
$$
t\mapsto 1-\left(\frac{t-1}{t}\right)^{t}- \left(\frac{t-1}{t}\right)^{t-1}
$$
is increasing on $[1, \infty)$.
\bibliographystyle{ims}
\bibliography{NP}

\end{document}